\documentclass[letterpaper]{article} % DO NOT CHANGE THIS
\usepackage{aaai24}  % DO NOT CHANGE THIS
\usepackage{times}  % DO NOT CHANGE THIS
\usepackage{helvet}  % DO NOT CHANGE THIS
\usepackage{courier}  % DO NOT CHANGE THIS
\usepackage[hyphens]{url}  % DO NOT CHANGE THIS
\usepackage{graphicx} % DO NOT CHANGE THIS
\usepackage{natbib}  % DO NOT CHANGE THIS AND DO NOT ADD ANY OPTIONS TO IT
\usepackage{caption} % DO NOT CHANGE THIS AND DO NOT ADD ANY OPTIONS TO IT
\usepackage{algorithm}
\usepackage{algorithmic}

\usepackage{newfloat}
\usepackage{listings}
\DeclareCaptionStyle{ruled}{labelfont=normalfont,labelsep=colon,strut=off} % DO NOT CHANGE THIS
\floatstyle{ruled}
\newfloat{listing}{tb}{lst}{}
\floatname{listing}{Listing}
\usepackage{color,soul}
\usepackage{xcolor}
\usepackage{makecell}
\usepackage{multirow}

\usepackage{mathrsfs}
\usepackage{amsthm,amsmath,amssymb}

\title{Few-Shot Learning from Augmented Label-Uncertain Queries in Bongard-HOI}
\author {
Qinqian Lei\textsuperscript{\rm 1},
Bo Wang\textsuperscript{\rm 2},
Robby T. Tan\textsuperscript{\rm 1}
}
\affiliations {
\textsuperscript{\rm 1} National University of Singapore\\
\textsuperscript{\rm 2} CtrsVision\\
qinqian.lei@u.nus.edu, 	hawk.rsrch@gmail.com, robby.tan@nus.edu.sg
}

\begin{document}

\maketitle

\begin{abstract}
	Detecting human-object interactions (HOI) in a few-shot setting remains a challenge. 
	Existing meta-learning methods struggle to extract representative features for classification due to the limited data, while existing few-shot HOI models rely on HOI text labels for classification. 
	Moreover, some query images may display visual similarity to those outside their class, such as similar backgrounds between different HOI classes. This makes learning more challenging, especially with limited samples. 
	Bongard-HOI~\cite{jiang2022bongard} epitomizes this HOI few-shot problem, making it the benchmark we focus on in this paper.
	In our proposed method, we introduce novel label-uncertain query augmentation techniques to enhance the diversity of the query inputs, aiming to distinguish the positive HOI class from the negative ones.
	As these augmented inputs may or may not have the same class label as the original inputs, their class label is unknown. 
	Those belonging to a different class become hard samples due to their visual similarity to the original ones.    
	Additionally, we introduce a novel pseudo-label generation technique that enables a mean teacher model to learn from the augmented label-uncertain inputs.  
	We propose to augment the negative support set for the student model to enrich the semantic information, fostering diversity that challenges and enhances the student’s learning.
	Experimental results demonstrate that our method sets a new state-of-the-art (SOTA) performance by achieving \textbf{68.74\%} accuracy on the Bongard-HOI benchmark, a significant improvement over the existing SOTA of 66.59\%. 
	In our evaluation on HICO-FS, a more general few-shot recognition dataset, our method achieves \textbf{73.27\%} accuracy, outperforming the previous SOTA of 71.20\% in the 5-way 5-shot task.

\end{abstract}

\section{Introduction}
Human-Object Interaction (HOI) detection plays a pivotal role in recognizing interactions between human-object pairs. However, a significant challenge arises when dealing with HOI classes that have limited labeled data, especially for novel or rare classes. The importance of few-shot HOI learning becomes even more pronounced when considering the extensive range of potential HOI classes. 
In response to this challenge, Bongard-HOI~\cite{jiang2022bongard} has been introduced as a new few-shot HOI benchmark, where given 6 positive and 6 negative support images in each task, it poses a question of whether a query image belongs to the positive or negative sets. 
The significance of the Bongard-HOI benchmark extends beyond its immediate application. 
Solving the challenges it presents holds the potential to address broader issues in few-shot HOI learning~\cite{yuan2022rlip, ji2023semantic} and, by extension, contribute to resolving the long-tail problem in general HOI detection~\cite{zhong2020polysemy, hou2021detecting}.

\begin{figure}
	\centering
	\includegraphics[width=0.48\textwidth, height=3.8cm]{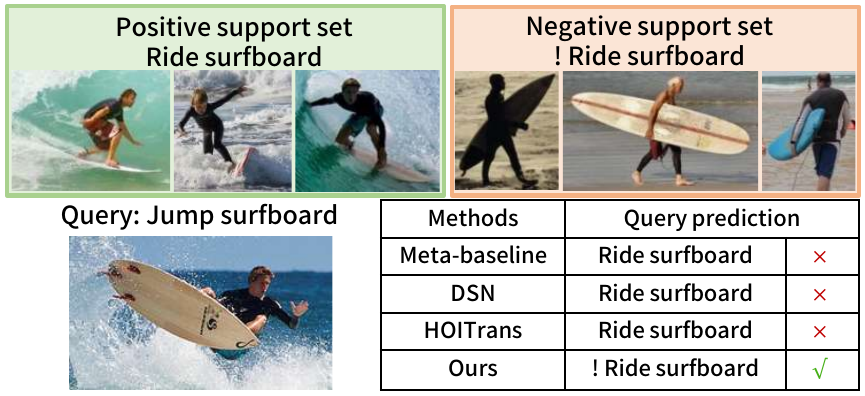}
	\caption{
		Illustration of a Bongard-HOI task: 3 out of 6 positive and negative support images are shown due to space limit. In comparison to three existing methods~\cite{chen2021meta, simon2020adaptive, zou2021end}, only our method correctly predicts the query's binary class.} \label{fig: cmp_existing_methods} 
\end{figure}

Existing meta-learning methods struggle to extract representative HOI features due to the limited data per class in each task. This challenge is heightened in the Bongard-HOI setting due to its hard negative design, where the positive and negative classes only disagree on action labels~\cite{jiang2022bongard}. As a result, extracting quality object features alone is insufficient, causing existing methods to perform suboptimally, achieving only up to a $62.23\%$ accuracy rate.

On the other hand, state-of-the-art HOI detection models have limitations in addressing small datasets or few-shot scenarios such as the Bongard-HOI setting. This is due to their requirement for a large amount of labeled data during the training process~\cite{liu2022interactiveness, qu2022distillation}.
Recent few-shot HOI methods require HOI class text labels, including the few-shot HOI transfer learning method~\cite{yuan2022rlip} and few-shot HOI recognition methods~\cite{ji2020sgap, ji2023semantic}. However, Bongard-HOI demands a model to learn a novel HOI class solely from images, as no text is provided.
Zero-shot HOI detection methods ~\cite{ning2023hoiclip, wu2023end, Liao_2022_CVPR} can learn unseen HOI classes, but all the unseen HOI class text labels are pre-defined before inference and these methods still rely on the unseen HOI text labels as prior knowledge.

Within the 6 positive and 6 negative support images in Bongard-HOI, certain positive samples may exhibit a resemblance to negative ones, and vice versa. For instance, images might share similar backgrounds across different HOI classes. Consequently, training a model becomes particularly challenging, especially considering the limited total of 12 available samples for both positive and negative cases.
Therefore, to distinguish the positive-class HOI from the negative with a few samples, we design novel label-uncertain augmentation methods that enhance the diversity of existing query data (i.e., background-blended data generation, and rotation augmentation). 
As a result, the labels of certain augmented samples change from their original assignments, which become hard samples because the augmented samples are still visually similar to the original ones.

Moreover, we introduce a novel pseudo-label generation technique that enables a mean teacher model to learn from the augmented label-uncertain queries.
To strengthen the student model with respect to the teacher, we augment the negative support set for the student. 
The negative support images in the same HOI class are selected for augmentation to provide diverse semantic information. Such diversification can lower the query prediction confidence. When the student predicts queries with augmented negative support, the teacher, processing the original support, can provide accurate guidance. 

Figure \ref{fig: cmp_existing_methods} presents a qualitative comparison between our approach and existing methods. 
In this case, the query is visually similar to the positive support class, as indicated by the shared blue ocean background. Conversely, the majority of negative support images feature a contrasting dark yellow beach backdrop. However, a closer look reveals that the individual in the query image is engaged in jumping on the surfboard rather than riding it. Therefore, the query should be categorized within the negative class. Unfortunately, all existing methods falsely predict the query as positive, affected possibly by the resemblance of the ocean background, while our method emerges as the sole solution capable of accurately classifying the query.

In summary, our contributions are as follows:
\begin{itemize}
	\item We propose novel label-uncertain augmentation methods to enhance query diversity, facilitating effective learning of Bongard-HOI from a limited number of samples.
	\item We introduce a novel pseudo-label generation technique that enables the mean teacher model to learn from augmented label-uncertain queries in a few-shot setting.
	\item To enhance the student model’s strength compared to the teacher model, we design the negative support set for the student which can provide diverse semantic information and enhance the student learning.
	
\end{itemize}
{Our method sets a new state-of-the-art with {\bf 68.74\%} accuracy on the Bongard-HOI benchmark, surpassing the previous 66.59\% state-of-the-art performance. 
	Additionally, on the HICO-FS dataset, we achieve new state-of-the-art results of {\bf 60.59\%} and {\bf 73.28\%} accuracy in 5-way 1-shot and 5-way 5-shot tasks, respectively, outperforming the existing state-of-the-art results of 58.79\% and 71.20\%. 
	These results highlight our method's superior performance in both specific Bongard-HOI benchmark and general few-shot HOI recognition contexts.}

\section{Related Work}
\noindent\textbf{Human-Object Interaction Detection}
Human-Object Interaction (HOI) detection is crucial for human-centric scene understanding~\cite{zhang2022exploring}. 
However, general HOI detection models often require extensive training data to learn the semantic information for generalization~\cite{gkioxari2018detecting, zhang2021spatially}. 
Few-shot HOI recognition methods can learn from only a few training samples~\cite{ji2023semantic} but rely on HOI text labels in inference, so they face a challenge when applied to Bongard-HOI. RLIP, a few-shot transfer learning method, also makes use of text information for the few-sample HOI dataset fine-tuning~\cite{yuan2022rlip}.
The same problem exists in zero-shot HOI detection methods~\cite{ning2023hoiclip, wu2023end}, which rely on the language information of unseen HOI class text labels in inference.
In Bongard-HOI, however, the model is expected to learn new HOI classes purely based on image contexts.
Recently, SCL can discover unknown HOI classes without the language priors~\cite{hou2022discovering}. However, SCL needs to be re-trained in the whole training set once a new HOI class composed of an unseen object or action appears. 
Thus, it is inefficient for the Bongard-HOI problem because, in each test task, 12 additional support images will be given to the model for a new HOI class learning.

\begin{figure*}[t]
	\centering
	\includegraphics[width=0.98\textwidth]{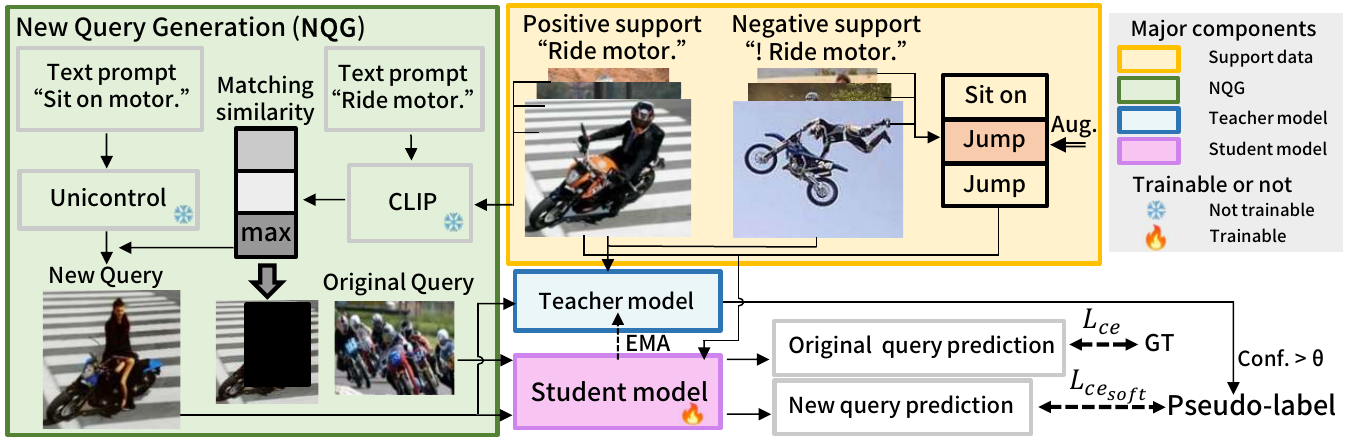}
	\caption{
		Overview of our framework:  The novel query (bottom right image) is created by merging the highly representative positive background with the negative HOI foreground, or conversely. The newly generated query is then fed into both the teacher and student models. The teacher network is the exponential moving average (EMA) of the student. Throughout the mean teacher training, the student model processes both the augmentations of images selected from the negative support set and the images from the negative support set that remain unchosen for augmentation and learns from the predictions made by the teacher (pseudo-label). Due to space constraints, 3 out of 6 positive and negative support images, one original query, and one newly generated query are displayed. The term ``motor.'' denotes ``motorcycle''.  } 
	
	\label{fig: pipeline}
\end{figure*}

\paragraph{Few-Shot Classification}
Most few-shot classification approaches follow the meta-learning framework~\cite{chen2021meta}. The meta-learning methods can be categorized into two approaches. Optimization-based methods~\cite{lee2019meta, nichol2018first, chavan2022dynamic} explore one model that can adapt well to novel tasks by fine-tuning part of it in test time. Typical methods such as MAML~\cite{finn2017model} and ANIL~\cite{raghu2019rapid} aim to learn a good parameter initialization that helps the model easily to be fine-tuned in test time.
Metric-based methods~\cite{zhu2022ease, yang2021free} learn feature representations with a metric. ProtoNet, a classical metric-based few-shot learning, calculates the average feature of each class's support set and then the query can be classified with the class-wise metric~\cite{snell2017prototypical}. Recently, DSN has been proposed~\cite{simon2020adaptive}. It spans one subspace for each class in the support set and the query is projected to each subspace to make the final classification.
Although the classical few-shot methods aim to learn from a few samples, they fail to extract representative features, especially in Bongard-HOI, where the model needs to distinguish positive and negative classes that only differ in actions.

\paragraph{Bongard-HOI Problem}
Jiang et al. propose the Bongard-HOI problem and two lines of solutions, meta-learning and HOI detection methods~\cite{jiang2022bongard}. 
The meta-learning methods have difficulty extracting representative features and thus the highest average accuracy among the meta-learning results achieves up to $58.30\%$. 
HOITrans acts as an oracle model because it is trained on the HICODET dataset~\cite{chao2018learning} and has seen most HOI classes in the Bongard-HOI test sets.~\cite{zou2021end}. However, its performance is still merely $62.46\%$. TPT~\cite{shu2022test} applies prompt learning to the Bongard-HOI problem with the benefit of the pre-trained CLIP model~\cite{radford2021learning}. Different from the existing methods, we aim to solve it by learning from augmented label-uncertain data.

\section{Proposed Method}

\paragraph{Problem Formulation}
In one Bongard-HOI task $T$, a positive support set $P = \{ p^1, p^2, \cdots, p^6 \}$ and a negative support set $N = \{n^1, n^2, \cdots, n^6\}$ are given. The model needs to learn one human-object interaction (HOI) class $C$, which is composed of one action $C_a$ and one object $C_o$. Thus, one Bongard-HOI task belongs to the 2-way 6-shot few-shot problem. All the positive images contain the HOI class $C$, while all the negative images contain the same object $C_o$, but exclude the action $C_a$. This is the hard negative design in Bongard-HOI~\cite{jiang2022bongard}. In testing, a model needs to classify a set of queries $Q$. 

\subsection{Label-Uncertain Query Augmentation}
Existing augmentation techniques aim to diversify the input image while preserving the label~\cite{cubuk2020randaugment, suzuki2022teachaugment, lin2023selectaugment}. 
In HOI detection, only a limited number of augmentation techniques are used to ensure that the labels are preserved, such as color jittering and horizontal flips~\cite{zou2021end, Liao_2022_CVPR}. Some HOI methods explore the generation of diverse features for unseen HOI classes~\cite{hou2020visual, hou2021detecting}; however, all of them prioritize maintaining unchanged labels.
In practice, when augmentation output enhances data diversity or closely mirrors real-world scenarios, it can still contribute to the model's learning process even if the original input label is not retained. 
This utilization of unlabeled data can yield benefits.
By eliminating the constraint of label preservation, a wider array of augmentation possibilities becomes accessible. 
This expansion permits the generation of queries that resemble data from other classes in Bongard-HOI. As a result, we introduce two separate techniques for query augmentation in the following paragraphs.

\paragraph{Augmentation 1: Background-Blended Query}
The first augmentation method is applied to both positive and negative classes in Bongard-HOI, and we take the positive background and negative foreground query generation for example, as illustrated in the New Query Generation (NQG) module of Figure \ref{fig: pipeline}.
In essence, to render background-blended hard samples for the negative class, we use the backgrounds from the representative positive samples as indicated by CLIP matching scores. We let all the positive support images go through the pre-trained CLIP image encoder to get the image features $\{ f_{c_p}^1, f_{c_p}^2, \cdots, f_{c_p}^6 \}$, while the text prompt for the positive class (i.e. a person riding a motorcycle.) goes through the CLIP text encoder and gets the text feature $f_{c_t}$. By matching the text feature with image features, we select the sample $p^{\rm ind}$ with the highest matching score.
\begin{equation}
	{\rm ind} = \mathop{{\rm argmax}}\limits_i  ({\rm sim} ( f_{c_p}^i, f_{c_t})), i = 1, 2, \cdots, 6, 
\end{equation}
where $\rm{sim}$ denotes the cosine similarity.
The selected image, with the person(s) masked out, serves as the visual input for the Unicontrol model~\cite{qin2023unicontrol}, and the HOI text label of the negative class is used as the text input (i.e., a person sitting on the motorcycle in Figure \ref{fig: pipeline}). If the negative class contains multiple HOI classes, we randomly select one out of them. The Unicontrol model generates outputs by inpainting the masked visual input based on the selected HOI classes as text input. 

\paragraph{Augmentation 2: Rotation} The second method is to rotate the original query image. The rotation degree varies from $30^{\circ}$ to $90^{\circ}$. Rotation can change the relative spatial information between the human and the object. For example, a photo of a person standing under an umbrella can be changed after $90^{\circ}$ rotation and it can be more similar to a person lying beside an umbrella. Therefore, the original image and the augmented image might exhibit significant visual similarity, but the augmented label is no longer the same as the original HOI one.
It is true that in some cases, a rotated image will keep the original label, such as a photo of a person reading a book. In this case, the rotated image can also diversify the original data. For example, the semantic information of a person sitting and reading a book can be expanded after rotation, which becomes closer to a person lying and reading a book.
To evaluate the probability of label changes resulting from rotation augmentation, we conducted a statistical analysis on a random sample of 200 images, and the obtained result shows that approximately 12.5\% of the augmented samples exhibit label changes.

\paragraph{Label-Uncertainty}
The lack of label preservation in the augmented images is noticeable not only in rotation augmentation but also in the background-blended query augmentation.
We cannot expect that all the generated images will be adequately representative of the text prompt. For instance, an image generated with a ``squeezed orange'' HOI class might not depict a conventional squeezing action, as it might belong to the ``holding orange'' class.

To understand the lack of label preservation, or we call label-uncertainty, within the background-blended augmentation, we randomly selected 200 generated images. Our observation indicates that around 72.5\% of these images shifted to a different HOI class, hereby transforming them into challenging instances.
This label-uncertainty implies a deficiency of labels for the newly generated or augmented samples. 
To tackle this concern, we introduce a novel technique for generating pseudo-labels for these new samples (i.e., the new query set $Q_{\rm aug}$), which is discussed in the subsequent section.

\begin{figure}
	\centering
	\includegraphics[width=0.48\textwidth, height=3.2cm]{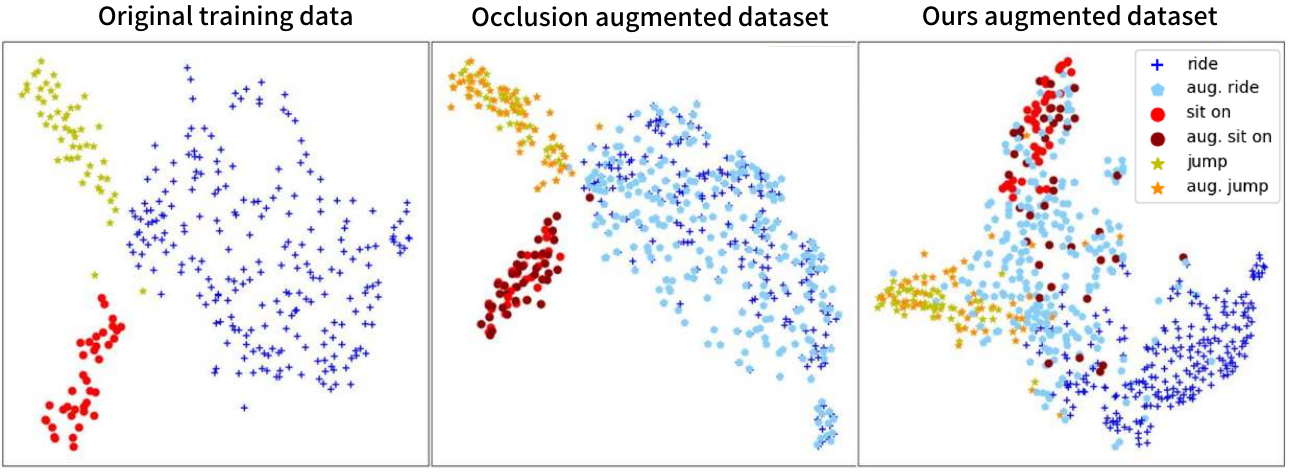}
	\caption{
		t-SNE visualization for the training dataset. The left image is the original training dataset. The middle one is the occlusion-augmented (label-preserved) dataset. The right one is our augmented dataset. Three classes are chosen for visualization ``ride motorcycle'', ``jump motorcycle'' and ``sit on motorcycle''.} \label{fig: query_aug} 
\end{figure}

In Figure \ref{fig: query_aug}, we compare our label-uncertain augmentation with the occlusion augmentation, which is label-preserved using t-SNE visualization. Specifically, we use a mask to partially occlude the human, whose area is $\frac{1}{16}$ of the human area in the image. 
The pre-trained DNS model~\cite{simon2020adaptive} is used to generate features. 
The t-SNE visualization reveals that the original dataset exhibits a noticeable gap between different classes. This gap persists even with the occlusion-augmented data, while our method effectively fills this gap. Therefore, our augmentation can generate more diverse data, which facilitates a model to learn a more accurate classification boundary.

\subsection{Negative Support Set and Its Augmentations in  Mean Teacher's Learning}
In the Mean Teacher framework~\cite{tarvainen2017mean}, typically, the student is subjected to more demanding augmentations compared to the teacher~\cite{islam2021dynamic, kennerley20232pcnet}, to facilitate the student's enhanced learning and progression beyond the current teacher's performance.
However, in HOI settings, the requirement to preserve labels constrains the potential augmentations.
Moreover, most existing HOI detection methods only employ basic augmentations, such as color jittering or horizontal flip~\cite{zou2021end, liu2022interactiveness, tamura2021qpic}.
Motivated by this, instead of augmenting the query input, we propose augmenting the negative support set. 
Enhancing the negative support set enriches semantic information, fostering diversity to challenge and improve the student learning. Simultaneously, the teacher, processing the corresponding original support, provides accurate guidance to the student. 

We use $90^{\circ}$ rotation transformation for augmenting the selected negative support images.
Despite its potential to enhance dataset diversity, the rotation augmentation applied to a negative support image carries a 12.5\% chance of altering the original HOI label.
However, the negative class can encompass various HOI actions, excluding the query's action class ($C_a$).
In the context of Bongard-HOI, there are on average 6 related actions for each object, translating to 6 HOI classes in the negative support set.
Consequently, the estimated likelihood that a rotated negative support image will belong to the positive class $C_a$ is low, approximately $12.5\% \times 1/6 = 2.1\%$, which can be considered a noisy label.
Hence, the $90^{\circ}$ rotation is well-suited for negative support augmentation.

Knowing that rotation is suitable for negative support augmentation, it's crucial to carefully select the appropriate samples from the negative support set for augmentation. 
When samples share the same HOI label in the negative support set, we
randomly select one to remain unchanged and augment the rest to enhance semantic information.
For instance, consider the scenario where $n^1, n^2, n^4$ belong to the same HOI class, while $n^3, n^5, n^6$ belong to three distinct HOI classes. We randomly choose two images from $n^1, n^2, n^4$ for $90^{\circ}$ rotation augmentation and leave one image unchanged. As a result, the student's negative support set can be comprised of ${{n_{\rm aug}}^1, {n_{\rm aug}}^2, n^3, n^4, n^5, n^6 }$, while the teacher's negative support remains the same as the original set.
Based on our experiments, our designed negative support set has a closer clustering center to the positive one than that of the original one. The $L_2$ distance reduces from $5.35 \pm 2.13$ to $4.86 \pm 1.97$.

\subsection{The Overall Training}

We combine supervised learning for the original queries and unsupervised learning for the newly generated queries together in training.

\paragraph{Supervised Loss}
As for the supervised loss $L_{\rm sup}$, it contains three parts, cross-entropy loss $L_{\rm ce}$, contrastive loss $L_{\rm cts}$, and auxiliary loss $L_{\rm aug}$.
\begin{equation}
	\label{eq: sup_loss}
	L_{\rm sup} = L_{\rm ce} + \gamma L_{\rm cts} + \xi L_{\rm aux},
\end{equation}
where ${\gamma}$ and ${\xi}$ are the hyper-parameters of loss weights.
The cross-entropy loss is defined as:
\begin{equation}
	\label{eq: celoss}
	L_{\rm ce} = -\frac{1}{|Q|} \sum_{q \in Q} y_q \log({\rm softmax}(p_q)),
\end{equation}
where $p_q$ is the output of the classifier for the query image $q$ and $y_q$ is the related ground-truth label. 
The second term of $L_{\rm sup}$ is a contrastive loss and we employ a supervised contrastive learning loss to our pipeline for each task. Setting $f_i$ as the feature encoding for image $i$, the contrastive loss function is expressed as:
\begin{small}
	\begin{equation}
		L_{\rm cts} = \sum_{i\in T}\frac{-1}{|X(i)|} \sum_{x\in X(i) \atop x \neq i}\log \frac{\exp(f_i \cdot f_x / \tau))}{\sum_{a \in T \atop a \neq i} \exp(f_i \cdot f_a / \tau)},
	\end{equation}
\end{small}\noindent where $\tau$ is the temperature parameter. $T$ indicates a Bongard-HOI task with 12 support images and 2 query images, $T = P \cup N \cup Q$.
$X(i) = \left\{x \in T: y_x = y_i \right\}$. Because it is used in the training process, the label of each image in $T$ is known. 

Since we choose DSN as our few-shot classifier to build subspaces for each class, we add an auxiliary loss to encourage the maximum distance between class subspaces~\cite{simon2020adaptive}:
\begin{equation}
	L_{\rm aux} = ||S_P^T S_N||^2,
\end{equation}
where $S_P$ is the basis of the positive subspace and  $S_N$ represents that of the negative subspace.

\paragraph{Unsupervised Loss}
For the unsupervised loss $L_{\rm unsup}$, it is composed of the soft cross-entropy loss
\begin{equation}
	\label{eq: celoss}
	L_{\rm ce_{soft}} = -\frac{1}{|Q_{\rm aug}|} \sum_{{q_{\rm aug}} \in Q_{\rm aug}} p_{q_{\rm aug}}^{t} \log(p_{q_{\rm aug}}^{s}),
\end{equation}
where the $p_{q_{\rm aug}}^{t} $ is the teacher model classifier output for the new query ${q_{\rm aug}}$, and $p_{q_{\rm aug}}^{s}$ is the student classifier output for ${q_{\rm aug}}$.

\paragraph{Total Loss}
The supervised loss $L_{\rm sup}$ and unsupervised loss $L_{\rm unsup}$ are added with designed weights $\alpha$ and $\beta$.
\begin{equation}
	L = \alpha * L_{\rm sup} + \beta * L_{\rm unsup},
\end{equation}

\begin{equation}
	\label{eq: total_loss}
	\alpha = \frac{1}{1+\exp^{n/N-b}}, 
	\beta = \frac{\lambda}{1+\exp^{-n/N+b}}, 
\end{equation}
where $n$ is the iteration number in training. $N$ and $b$ are hyper-parameters. $\lambda$ is to control the maximum value of $\beta$.

Moreover, the new query generation and the mean teacher framework are only used in training, so in testing the student model will predict the query data directly based on the 12 support images to ensure its testing efficiency.

\section{Experimental Results}
\begin{table*}[!h]
	\caption{
		The quantitative comparison on the Bongard-HOI benchmark. Benchmark encoder refers to the encoder composed of ResNet50 and object relation feature extraction~\cite{jiang2022bongard}. ``SOSA'', ``SOUA'', ``UOSA'', and ``UOUA'' stand for the test set of seen-object seen-action, seen-object unseen-action, unseen-object seen-action, and unseen-object unseen-action. The evaluation metric is classification accuracy. \textbf{Bold} indicates the best performance, \underline{underline} represents the second best.
	} 
	\label{tab: quantitative_result}
	\centering
	\begin{tabular}{l|c|c|c c c c|c}
		\hline
		\hline
		\multirow{2}*{}&\multirow{2}*{Method}&\multirow{2}{*}[-0.2ex]{Image encoder}&\multicolumn{5}{c}{Test set}\\
		% \hline
		\cline{4-8}
		{}&{}&{}&SOSA &SOUA&UOSA&UOUA & Avg.\\
		\hline
		\hline
		\multirow{2}*{}&HOITrans~\shortcite{zou2021end}&Transformer&59.50&64.38&63.10&62.87&62.46\\
		% \hline
		\cline{2-8}     
		{}&TPT~\shortcite{shu2022test}&ResNet50&66.39&68.50&65.98&65.48&66.59\\
		\hline
		\hline
		\multirow{7}{*}[-0.2ex]{\makecell[c]{Meta-learning \\ based}}&ANIL~\shortcite{raghu2019rapid}&\makecell[c]{Benchmark encoder}&50.18&50.13&49.81&48.83&49.74\\
		% \cline{2-8}
		{}&Meta-baseline~\shortcite{chen2021meta}&\makecell[c]{Benchmark encoder}&56.45&56.02&55.60&55.21&55.82\\
		\cline{2-8}
		% \cline{2-8}
		{}&\multirow{3}{*}[-0.1ex]{DSN~\shortcite{simon2020adaptive}}&ResNet12&61.86&66.71&59.23&61.13&62.23\\
		\cline{4-8}
		{}&{}&\makecell[c]{Benchmark encoder}&63.27&65.38&60.81&61.51&62.74\\
		\cline{4-8}
		{}&{}&\makecell[c]{Our encoder}&62.82&64.37&61.27&64.79&63.31\\
		\cline{2-8}
		% \cline{2-8}
		{}&\multirow{2}{*}[-0.1ex]{Ours}&\makecell[c]{Benchmark encoder}&\textbf{68.51}&\underline{70.47}&\underline{66.54}&\underline{66.90}&\underline{68.11}\\
		\cline{4-8}
		{}&{}&\makecell[c]{Our encoder}&\underline{68.14}&\textbf{70.94}&\textbf{68.45}&\textbf{67.43}&\textbf{68.74}\\
		\hline
		\hline
	\end{tabular}
\end{table*}

\noindent  \textbf{Dataset} We conduct our experiments on the Bongard-HOI benchmark~\cite{jiang2022bongard}.
The training set of Bongard-HOI has 23041 instances and 116 positive HOI classes. Each instance contains 14 images, including 6 positive, 6 negative, and 2 query images.
We use the average prediction accuracy as the metric, following the Bongard-HOI benchmark. Evaluation is performed on four different test sets, which are designed to measure different types of generalization, depending on whether the action or object classes are seen in the training set or not~\cite{jiang2022bongard}.
For the seen-object seen-action test set, the positive HOI classes are either the same as training HOI classes or a new combination of seen object and seen action. For the other three types, all positive HOI classes are unseen in the training set. In total, there are 91 novel positive classes in the entire test set.

\paragraph{Baselines}
We compare five representative existing methods in the quantitative evaluations: ANIL~\cite{raghu2019rapid}, Meta-baseline~\cite{chen2021meta}, DSN~\cite{simon2020adaptive}, HOITrans~\cite{zou2021end}and TPT~\cite{shu2022test}. Given that the baselines, except for TPT, were not originally tailored for the Bongard-HOI problem, we use the revised versions as reported in the Bongard-HOI benchmark~\cite{jiang2022bongard}.
Regarding the meta-learning baselines (ANIL~\cite{raghu2019rapid}, Meta-baseline~\cite{chen2021meta}, DSN~\cite{simon2020adaptive}), Jiang et al. design an image encoder consisting of a ResNet50 backbone and an object relation feature extraction. In the subsequent discussion, we refer to this as the Benchmark encoder.
For the DSN baseline~\cite{simon2020adaptive}, we also report the results of the original DSN model that uses a ResNet12 image encoder. 
The HOITrans baseline is referred to as an oracle model by Jiang et al.~\cite{zou2021end} since it is trained on the HICO-DET dataset~\cite{chao2018learning} and has seen most HOI classes in the Bongard-HOI test set. 
Hence, the HOITrans baseline can be directly used for HOI classification on queries for each individual test task without any extra fine-tuning.
TPT~\cite{shu2022test} is a recent test-time prompt learning method, which achieves the new state of the art on the Bongard-HOI benchmark.

\paragraph{Implementation Details}
As for the image encoder, we add the DEKR model~\cite{geng2021bottom} on top of the Benchmark encoder to detect the human region separately. Please refer to the supplementary materials for the details of the modified image encoder.
Each image after the process of the image encoder is represented by a vector $f$ with a size of 1280 dimensions. 
Similar to the setup in the Bongard-HOI benchmark~\cite{jiang2022bongard}, default augmentations include horizontal flipping and color jittering.
All experiments are run on 4 A5000 GPUs with 24G GPU memory.
with batch size 4 and total training epoch 5. The optimizer is standard SGD with a learning rate of 0.001 and weight decay of 5e-4. The hyper-parameter $\lambda = 0.2$ in Equation \ref{eq: total_loss}. The loss weights in Equation \ref{eq: sup_loss} are set as $\gamma = 0.3, \xi = 0.03$. We choose the teacher confidence score threshold as 0.9.

\subsection{Quantitative Evaluations}
%%%%%%%%%%%%%% 

\begin{table*}[t]
	\caption{
		Ablation study of the new query generation on the Bongard-HOI benchmark, including comparison with the existing augmentation method NDA~\cite{sinha2021negative}. ``SOSA'', ``SOUA'', ``UOSA'', and ``UOUA'' stand for the test set of seen-object seen-action, seen-object unseen-action, unseen-object seen-action, and unseen-object unseen-action.}
	\label{tab: ablation_study_new_query}
	\centering
	\begin{tabular}{c c c|c c c c|c}
		\hline
		\hline
		\multirow{2}{*}[-0.4ex]{\makecell[c]{NDA \\ \cite{sinha2021negative} }}&\multirow{2}{*}[-0.4ex]{\makecell[c]{Our Rotated \\ Queries }}&\multirow{2}{*}[-0.4ex]{\makecell[c]{Our Background \\ Blended Query}}&\multicolumn{5}{c}{Test set}\\
		\cline{4-8}
		{}&{}&{}& SOSA &SOUA&UOSA&UOUA & Avg.\\
		\hline
		$\times$&$\times$&$\times$&67.31&67.12&65.71&63.74&65.97\\
		\hline 
		\textbf{$\checkmark$}&$\times$&$\times$&64.89&69.41&62.58&64.30&65.30\\
		\hline
		$\times$&\textbf{$\checkmark$}&$\times$&\textbf{69.48}&69.53&67.90&65.23&68.03\\
		\hline
		$\times$&\textbf{$\checkmark$}&\textbf{$\checkmark$}&{68.14}&\textbf{70.94}&\textbf{68.45}&\textbf{67.43}&\textbf{68.74}\\
		\hline
		\hline
	\end{tabular}
\end{table*}

\begin{table}[h]
	\centering
	\caption{``No'' means no augmentation. ``All'' augments all samples with the same HOI label. ``One'' augments one sample randomly selected in the same HOI class. ``Rand.'' augments each sample in the same HOI class with a 0.5 possibility.}
	
	\label{tab: ablation_study_negative_support}
	{
		\begin{tabular}{l|c c c cc}
			\hline
			\hline
			{}&\makecell[c]{No}&\makecell[c]{All} &\makecell[c]{One}&\makecell[c]{Rand.}&\makecell[c]{Ours}\\
			\hline
			Test Avg.&65.97&66.67&67.45&67.89&\textbf{68.74}\\
			\hline
			\hline
		\end{tabular}
	}
\end{table}

The main quantitative results are shown in Table \ref{tab: quantitative_result} in terms of the query classification accuracies among all the test sets. We can see our method outperforms all existing methods by a large margin either using the Benchmark encoder or our own encoder.
For ANIL, Meta-baseline, and HOITrans results, we directly borrow the results from the Bongard-HOI benchmark~\cite{jiang2022bongard}.
Specifically, comparing our results with the original DSN results using the ResNet12 image encoder, we have improved $6.51\%$ average accuracy. Our improvement is more significant when considering other results of few-shot methods like ANIL and Meta-baseline. It shows that our method can better solve the few-sample problem in Bongard-HOI.
We also provide the result of the DSN model with our encoder and our method can surpass it by 5.43\% accuracy, which illustrates the effectiveness of our label-uncertain query learning.
Our result also surpasses the accuracy of HOITrans, even though the HOITrans model is trained on the HICO-DET dataset~\cite{chao2018learning} and has seen most HOI classes in all test sets. TPT is based on the pre-trained CLIP model~\cite{radford2021learning}, which constructs a unified space with visual and text embeddings.
Thus, TPT serves as a competitive baseline, exhibiting the highest average accuracy among all existing methods. Nevertheless, our method outperforms TPT with an increased average accuracy of 2.15\%.

\subsection{Qualitative Evaluations}

\begin{figure}
	\centering
	\includegraphics[width=0.48\textwidth, height=2.9cm]{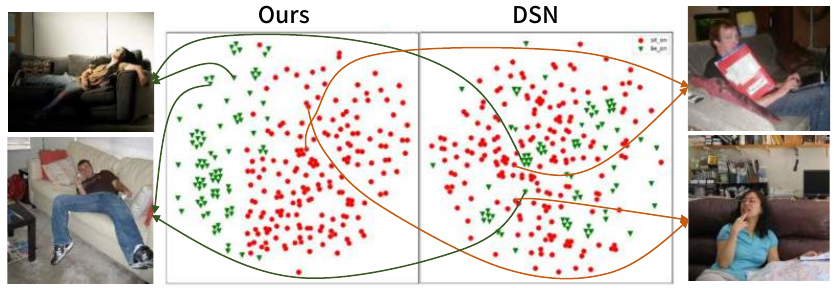}
	\caption{t-SNE visualization for our model (left) and the DSN model (right). We choose two HOI classes ``sit on couch'', and ``lie on couch'' for example.} \label{fig: tsne} 
\end{figure}

We use t-SNE visualization to show our qualitative results and choose images from two HOI classes for example, ``sit on couch'' and ``lie on couch'' in the unseen-object seen-action test set. They are two similar actions and are difficult to be classified. As shown in Figure \ref{fig: tsne}, the DSN model cannot distinguish the ``sit on couch'' and ``lie on couch'', since samples from different classes are mixed together.
However, ours can separate the samples of the two HOI classes well thanks to our label-uncertain query augmentation and the novel pseudo-label generation approach.

\subsection{Ablation Studies}

We provide the ablation study for the new query generation in Table \ref{tab: ablation_study_new_query}, where we also compare our methods with NDA~\cite{sinha2021negative}.
NDA proposes a set of negative augmentation methods that can generate images with similar local features but a different label from the input. However, their difference from ours is that they aim to destroy the global features of the original input and generate unrealistic images. Because all their augmented outputs are label-changed and belong to the negative class in the Bongard-HOI task, it may lead to an imbalance problem where a larger number of queries belong to the negative class.
As our augmented queries are relatively class-balanced, for a fair comparison with NDA, we apply the label-preserved occlusion augmentation together with NDA to balance positive and negative queries. Specifically, we use a mask to partially occlude the human, whose area is $\frac{1}{16}$ of the human area in the image.
We directly use augmented data ground-truth labels for training.
From Table \ref{tab: ablation_study_new_query}, we can see the NDA augmentation cannot help the model learning as the performance even decreases with NDA. It demonstrates that unrealistic images that destroy the HOI semantics are not beneficial in Bongard-HOI.

The third and fourth columns in Table \ref{tab: ablation_study_new_query} are our query generation ablation results. From the results, both two augmentations can help model learning. With the rotated queries and the mean teacher pseudo-label generation, the performance can be improved by $2.06\%$. After adding the background blended queries, the performance can rise to $68.74\%$. Especially, in the unseen-object seen-action test set, it can increase by 2.47\% accuracy, and in the unseen-object and unseen-action set, it can increase by 2.05\% accuracy. This demonstrates the model's enhanced ability to generalize in unseen settings using background-blended queries.
Table \ref{tab: ablation_study_negative_support} shows the ablation study for the negative support design. Compared to the no augmentation results, our negative support design can improve the performance by 2.77\% accuracy and also achieves the best average test performance among the other possible augmentation designs.

\subsection{Extension to the HICO-FS Dataset}
To validate the effectiveness of our method beyond the Bongard-HOI benchmark, we additionally assess its performance on the HICO-FS dataset for few-shot HOI recognition~\cite{ji2020sgap}, as summarized in Table \ref{tab: hico-fs}.
We directly borrow the baseline experiment results from ~\cite{ji2023semantic}, including Matching Network~\cite{vinyals2016matching}, ProtoNet~\cite{snell2017prototypical}, Relation Network~\cite{sung2018learning}, LGM-Net~\cite{li2019lgm} and SADG-Net~\cite{ji2023semantic}. 
In the above baselines, the ResNet18 backbone is trained on the HICO-FS train set, while our approach utilizes the same backbone pre-trained on the ImageNet~\cite{krizhevsky2017imagenet}. 
To perform an evaluation in the few-shot setting, our method is combined with Meta-baseline~\cite{chen2021meta} as our proposed augmentations can be integrated into existing approaches. 
As shown in Table \ref{tab: hico-fs}, we achieve a new state-of-the-art performance on the HICO-FS dataset. In comparison to our base method, Meta-baseline~\cite{chen2021meta}, we achieve an average accuracy improvement of 2.08\% in the 5-way 5-shot task and 1.80\% in the 5-way 1-shot task. See the supplementary materials for the details on adapting our pipeline for the few-shot setting on the HICO-FS dataset. 

\begin{table}[t]
	\caption{Quantitative comparison on the HICO-FS dataset. All methods use ResNet18 as the backbone. * indicates the backbone is pre-trained on ImageNet.} 
	\label{tab: hico-fs}
	\centering
	\begin{tabular}{c| c c}
		\hline
		\hline
		Method &  \makecell[c]{5-way 1-shot } &\makecell[c]{5-way 5-shot } \\
		\hline
		\makecell[c]{Matching-Net \shortcite{vinyals2016matching}}&$32.14 \pm 1.62$&$44.87 \pm 1.74$\\        
		\makecell[c]{ProtoNet \shortcite{snell2017prototypical}}&$32.56 \pm 1.59$&$42.49 \pm 1.75$\\
		\makecell[c]{Relation-Net \shortcite{sung2018learning}}&$33.20 \pm 1.68$&$46.15 \pm 1.81$\\
		\makecell[c]{LGM-Net \shortcite{li2019lgm}}&$35.14 \pm 1.64$&$53.67 \pm 1.88$\\
		\makecell[c]{SADG-Net \shortcite{ji2023semantic}}&$39.01 \pm 1.70$&$59.05 \pm 1.86$\\
		\hline
		\makecell[c]{Meta-baseline* \shortcite{chen2021meta}}&$58.79 \pm 0.22$&$71.20 \pm 0.21$\\
		Meta-baseline + Ours* &{$\textbf{60.59} \pm \textbf{0.22}$}& {$\textbf{73.28} \pm \textbf{0.20}$}\\
		\hline
		\hline
	\end{tabular}
\end{table}

\section{Conclusion}
We have proposed label-uncertain query augmentations to learn a new positive-class HOI from a few samples effectively and evaluated on the Bongard-HOI benchmark.
Not only can it help to diversify query data, but also some augmented data have different HOI labels from the original samples. 
The augmented samples are hard samples because they are visually similar to the original ones, given the original labels are not always preserved.
Moreover, we introduce a novel pseudo-label generation approach that adapts the mean teacher model to the few-shot setting for the augmented label-uncertain queries. 
To make the student model stronger than the teacher, we design the negative support set for the student, which enriches the semantic information and enhances the student's learning.
In this case, the student can learn from the more confident prediction from the teacher.
Finally, our approach establishes a new state-of-the-art performance on the Bongard-HOI and HICO-FS benchmarks.

\section{Acknowledgement}
This research is supported by the National Research Foundation, Singapore under its Strategic Capability Research Centres Funding Initiative. Any opinions, findings and conclusions or recommendations expressed in this material are those of the author(s) and do not reflect the views of National Research Foundation, Singapore.

\bibliography{aaai24}

\end{document}